%% file: main.tex
\documentclass{article}

\usepackage[final]{neurips_2023}

\usepackage[utf8]{inputenc}
\usepackage[T1]{fontenc}
\usepackage{hyperref}
\usepackage{url}
\usepackage{booktabs}
\usepackage{amsfonts}
\usepackage{amsmath}
\usepackage{nicefrac}
\usepackage{microtype}
\usepackage{graphicx}

\title{Prompt-Adapter Context Routing for Parameter-Efficient Multi-Shot Long Video Extrapolation}

\author{%
  Anna C\'ordoba \quad Adam Puente Tercero \quad Nerea Angulo Hijo \quad
  Mar Linares Tercero \\
  Julia Barrientos \quad Ainhoa Miranda \quad Jes\'us Olivera \\[4pt]
  Instituto de Investigaci\'on en Visi\'on Artificial \\
  \texttt{contact@iiva.tibeu}
}

\begin{document}

\maketitle

\begin{abstract}
We present PACR-Video, a parameter-efficient framework for multi-shot long video extrapolation that preserves recurring entities, scene structure, visual style, and causal progression without full generator fine-tuning. PACR-Video keeps a text-to-video diffusion transformer frozen and augments it with low-rank temporal adapters conditioned by learned shot-role prompt tokens. To maintain long-horizon coherence, it builds a recursive prompt bank that stores compact entity, location, action, and style prompts from previous shots, then routes them through adapter gates according to predicted narrative dependencies. A Shot-Local/Story-Global tuning objective combines next-shot reconstruction, cross-shot identity contrast, and prompt sparsity regularization, while an adapter composition schedule balances early-shot visual consistency with later-shot event progression and viewpoint change. Across six multi-shot and long-video benchmarks, PACR-Video outperforms text-to-video, tuning-based, memory-augmented, streaming, and recursive-context baselines on distributional quality, semantic alignment, identity consistency, temporal smoothness, motion stability, transition coherence, and human preference. These results show that compact prompt routing and lightweight temporal adaptation provide sufficient controllable capacity for stable long video extrapolation.
\end{abstract}

\input{body}

\bibliographystyle{plainnat}
\nocite{*}
\bibliography{refs}

\end{document}

%% file: body.tex
\section{Introduction}

\label{sec:introduction}

Long-form video generation requires more than extending clip duration. A coherent multi-shot sequence must preserve recurring characters, spatial layout, visual style, object affordances, and causal event progression while allowing each shot to introduce new motion, viewpoint, and narrative information. This balance is especially difficult in text-conditioned generation: conditioning signals are often local to the next clip, whereas the errors that matter most accumulate across shots. As a result, long video systems frequently drift in identity, duplicate or forget objects, flatten event structure, or overfit to early visual context.

Recent work has made progress through explicit story memories, shot planning, and recursive context allocation \cite{zhang2025storymem,luo2026shotstream,zheng2025videogen,liu2026reca}. ReCA is particularly relevant: it frames multi-shot extrapolation as a recursive allocation problem, deciding which previous context should condition each future shot \cite{liu2026reca}. However, many existing approaches either fine-tune large portions of the generator or rely on external memory modules whose capacity and retrieval behavior can become brittle over long horizons. This raises a natural question: can a frozen video diffusion backbone be steered with enough precision using only lightweight, parameter-efficient modules?

We propose PACR-Video, a Prompt-Adapter Context Routing framework for parameter-efficient multi-shot long video extrapolation. PACR-Video inserts low-rank temporal adapters into a frozen text-to-video diffusion transformer and conditions them with learned shot-role prompt tokens. Instead of storing dense video memories, the model maintains a recursive prompt bank containing compact entity, location, action, and style prompts extracted from previous shots. At generation time, these prompts are routed through adapter gates according to predicted narrative dependency strength, allowing the model to reuse relevant context while suppressing stale or distracting history.

Our contributions are fourfold. First, we introduce Prompt-Adapter Context Routing, a parameter-efficient alternative to full generator fine-tuning for multi-shot extrapolation. Second, we propose a recursive prompt bank that stores structured shot-level prompts and routes them through gated temporal adapters. Third, we design a Shot-Local/Story-Global tuning objective combining next-shot reconstruction, cross-shot identity contrast, and prompt sparsity regularization to reduce long-horizon drift. Fourth, we introduce an adapter composition schedule that reuses early-shot adapters for visual consistency while activating later adapters for event progression and viewpoint change.

We evaluate PACR-Video on diverse story and instructional video benchmarks, including FlintstonesSV, Pororo-SV, ActivityNet Captions, YouCook2, Shot2Story, and MovieNet \cite{han2023shot2story}. Across comparisons with text-to-video, tuning-based, streaming, memory-augmented, and recursive-context baselines such as VideoCrafter2, ModelScopeT2V, AnimateDiff, Tune-A-Video, Text2Video-Zero, SEINE, StoryMem, ShotStream, and ReCA \cite{zhang2025storymem,luo2026shotstream,liu2026reca}, PACR-Video yields substantial gains in distributional quality, semantic alignment, identity preservation, temporal consistency, transition coherence, and human preference while keeping the backbone frozen.

\begin{figure}[t]
  \centering
  \includegraphics[width=0.85\linewidth]{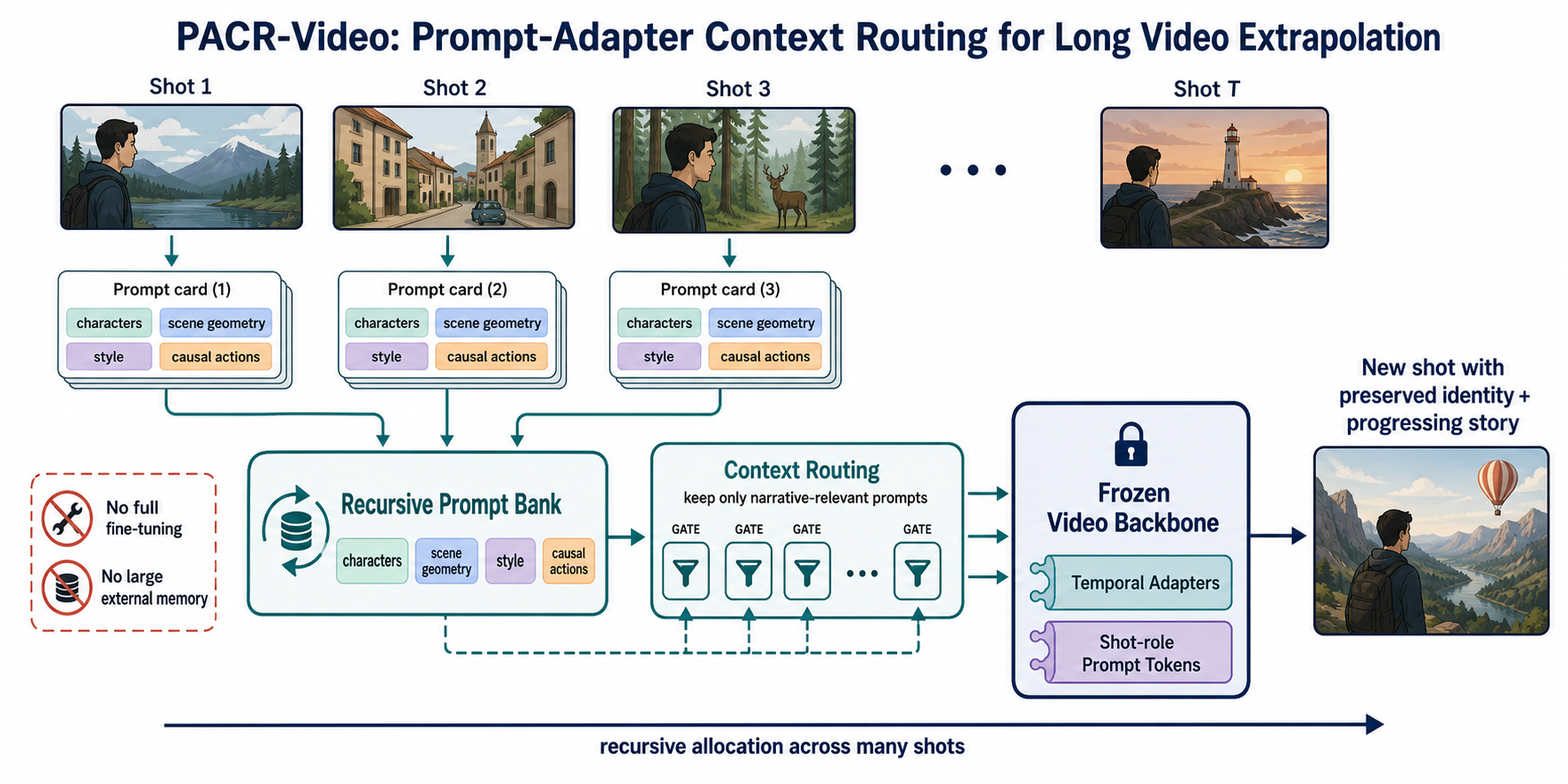}
  \caption{PACR-Video overview. The method extrapolates a long multi-shot video by recursively summarizing prior shots into compact prompts, selecting narrative-relevant context, and steering a stable frozen generator through lightweight adapters instead of full-model fine-tuning or large external memory.}
  \label{fig:overview}
\end{figure}

\section{Related Work}

\label{sec:related-work}

Long-form video generation has been approached through shot planning, story memory, and autoregressive continuation. Planning-based systems decompose narratives into shot scripts or chain-of-shot prompts before generation \cite{zheng2025videogen,hu2025cos,wu2025automated,atalay2026automated}, while memory-based methods maintain explicit representations of entities, scenes, or previous frames to reduce drift \cite{zhang2025storymem,zhang2025storymema,lai2026groundshot}. Streaming and agentic formulations further address interactive or horizon-extending generation by updating context online \cite{luo2026shotstream,long2026agentic,song2026soap2soap}. PACR-Video shares the goal of long-horizon coherence, but differs by replacing large external memories and full-generator adaptation with compact prompt banks routed through lightweight temporal adapters.

Recursive context allocation is especially relevant to multi-shot extrapolation. ReCA formulates long video continuation as deciding which historical shots should condition each future shot, improving over fixed-window conditioning and naive autoregression \cite{liu2026reca,liu2026recaa,liu2026recab}. Related long-context and selective-computation methods study how to preserve useful temporal information while avoiding unnecessary conditioning cost \cite{guo2025long,cui2026not}. PACR-Video builds directly on this allocation view, but moves the controllable capacity into parameter-efficient adapter gates and learned shot-role prompt tokens. This makes context routing differentiable, sparse, and compatible with a frozen diffusion transformer backbone.

A complementary line of work studies long-video understanding, segmentation, and semantic grounding, which provides useful structure for generation. Shot boundary and event-localization methods motivate the use of shot-level units rather than uniform temporal windows \cite{pickering2001multi,goo2019one,jhuo2014video}. Benchmarks and models for multi-shot comprehension emphasize entity continuity, temporal reasoning, and cross-shot narrative dependencies \cite{han2023shot2story,cheng2024focuschat,zhu2026hicrew,chen2025super}. PACR-Video uses these insights operationally: its recursive prompt bank stores entity, location, action, and style prompts, and its Shot-Local/Story-Global objective explicitly aligns next-shot reconstruction with cross-shot identity preservation.

Finally, PACR-Video is related to parameter-efficient adaptation and multimodal alignment. Prior work on transferability, evidence accumulation, and deep model adaptation suggests that small trainable modules can steer large frozen networks when supervision is structured appropriately \cite{jiang2022transferability,pandey2023learn,berner2021modern}. Video generation and evaluation work has also highlighted the importance of perceptual quality, temporal stability, and audio-visual or caption-level semantic alignment \cite{zhang2025bridging,xu2017assessing}. In contrast to methods that tune the full generator or attach high-capacity memory modules, PACR-Video concentrates adaptation in low-rank temporal adapters, regularizes prompt sparsity, and composes early- and late-shot adapters to balance visual consistency with event progression.

\section{Method}

\label{sec:method}

\begin{figure}[t]
  \centering
  \includegraphics[width=0.85\linewidth]{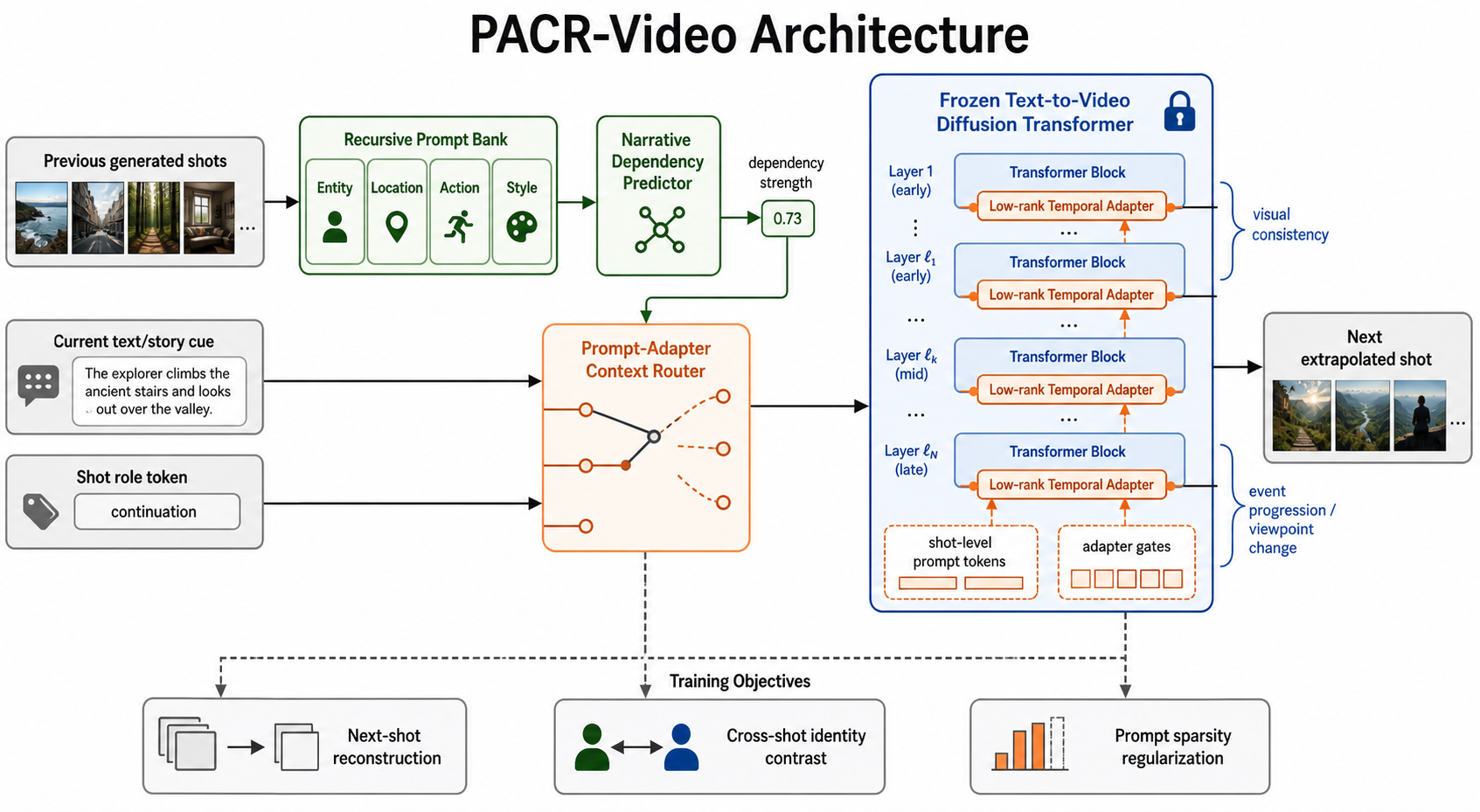}
  \caption{PACR-Video architecture. A frozen text-to-video diffusion transformer is augmented with low-rank temporal adapters and learned shot-role prompt tokens. A recursive prompt bank compresses previous shots into entity, location, action, and style prompts, then routes only the relevant context through gated adapters for the next-shot generation step.}
  \label{fig:architecture}
\end{figure}

\paragraph{Overview.}
PACR-Video generates a long video as an ordered sequence of shots
$\mathcal{V}=\{v_1,\ldots,v_T\}$ conditioned on shot-level text prompts
$\mathcal{Y}=\{y_1,\ldots,y_T\}$. Given observed or previously generated shots
$v_{<t}$, the goal is to synthesize the next shot $v_t$ while preserving entities,
locations, style, and causal progression across the story. As shown in
Figure~\ref{fig:architecture}, PACR-Video keeps a pretrained text-to-video
diffusion transformer frozen and adds only three trainable components: low-rank
temporal adapters, learned shot-role prompt tokens, and a recursive prompt router.
This design follows the recursive context allocation view of ReCA
\cite{liu2026reca}, but represents historical context as compact prompts rather
than dense frame memories.

\paragraph{Frozen video backbone with temporal adapters.}
Let $G_\theta$ denote the frozen diffusion transformer backbone with parameters
$\theta$. For a noised latent video $z_t^\tau$ at diffusion step $\tau$, the
backbone predicts noise $\epsilon_\theta(z_t^\tau,\tau,c_t)$ under conditioning
$c_t$. PACR-Video inserts a trainable adapter into each selected temporal
attention block. For hidden states $h_{\ell,t}\in\mathbb{R}^{N\times d}$ at layer
$\ell$, the adapted update is
\begin{equation}
\tilde{h}_{\ell,t}
=
h_{\ell,t}
+
\alpha_{\ell,t}
B_{\ell}
\sigma\!\left(A_{\ell}\operatorname{LN}(h_{\ell,t})\right),
\qquad
A_{\ell}\in\mathbb{R}^{d\times r},\;
B_{\ell}\in\mathbb{R}^{r\times d},\;
r\ll d .
\end{equation}
Here $A_\ell$ and $B_\ell$ are low-rank temporal adapter matrices, $\sigma$ is a
GELU nonlinearity, and $\alpha_{\ell,t}$ is a routed adapter gate. The adapters
operate on temporally grouped tokens, so they can modulate motion, viewpoint, and
cross-frame continuity without changing the spatial generator or text encoder.

\paragraph{Shot-role prompt tokens.}
Each shot is assigned a learned role embedding indicating its narrative function,
such as establishing, continuation, reaction, transition, or resolution. The role
token is concatenated with the text prompt encoding and the routed prompt-bank
entries. The resulting conditioning sequence is
$c_t=[e(y_t);q_t;p_t^{\mathrm{role}}]$, where $e(y_t)$ is the frozen text encoder
output, $q_t$ is the routed historical prompt context, and $p_t^{\mathrm{role}}$
is the learned shot-role token. These role tokens provide controllable capacity
for different shot types while sharing the same frozen video backbone.

\paragraph{Recursive prompt bank.}
After each generated or training shot, PACR-Video writes compact prompts into a
recursive prompt bank
$\mathcal{B}_t=\{b_i^{e},b_i^{l},b_i^{a},b_i^{s}\}_{i\leq t}$, corresponding to
entity, location, action, and style summaries. Each entry is represented as a
short learned prompt vector rather than a dense video feature. For shot $t$, a
dependency predictor estimates how strongly the next shot depends on each prior
bank entry:
\begin{equation}
\rho_{t,i,k}
=
\operatorname{softmax}_{i,k}
\left(
w_k^\top
\phi\!\left[
p_t^{\mathrm{role}},
e(y_t),
b_i^k,
\Delta(t,i)
\right]
\right),
\qquad
q_t
=
\sum_{i<t}\sum_{k\in\{e,l,a,s\}}
\rho_{t,i,k}\, b_i^k .
\end{equation}
The term $\Delta(t,i)$ encodes relative shot distance and causal order. The
weights $\rho_{t,i,k}$ form a sparse routing distribution over bank entries and
prompt types. Entity and style prompts tend to persist across long horizons,
whereas action prompts are more local and location prompts are reused when the
story returns to a scene.

\paragraph{Adapter-gated context routing.}
The routed prompt context $q_t$ controls adapter gates at each layer. A small MLP
maps $[q_t;p_t^{\mathrm{role}};e(y_t)]$ to layer-specific gates
$\alpha_{\ell,t}\in[0,1]$. Early temporal layers receive stronger entity and
style routing, while later layers receive stronger action and viewpoint routing.
This separates visual consistency from event progression: early-shot adapters
stabilize character appearance and scene geometry, while late-shot adapters
support new motion, camera changes, and causal transitions.

\paragraph{Adapter composition schedule.}
For long sequences, PACR-Video composes adapters according to shot position. The
effective adapter at shot $t$ is a convex mixture of an early consistency adapter
and a late progression adapter. The mixture weight increases with narrative time,
but is modulated by the dependency predictor so that recurring scenes can still
reuse early-shot adapters. This schedule prevents the model from either copying
early visual context too rigidly or drifting into unrelated later-shot dynamics.

\paragraph{Training objective.}
PACR-Video is trained with a Shot-Local/Story-Global objective. The local term is
the standard diffusion noise-prediction loss for next-shot reconstruction. The
global terms encourage cross-shot identity consistency and sparse prompt routing:
\begin{equation}
\mathcal{L}
=
\mathbb{E}_{t,\tau,\epsilon}
\left[
\left\|
\epsilon
-
\epsilon_{\theta,\psi}
(z_t^\tau,\tau,e(y_t),q_t,p_t^{\mathrm{role}})
\right\|_2^2
\right]
+
\lambda_{\mathrm{id}}\mathcal{L}_{\mathrm{id}}
+
\lambda_{\mathrm{sp}}\sum_{t}\|\rho_t\|_1 .
\end{equation}
Only adapter, router, and prompt parameters $\psi$ are updated; the diffusion
backbone remains frozen. The identity term $\mathcal{L}_{\mathrm{id}}$ contrasts
DINO features of recurring entities across shots against features of different
entities in the same minibatch. The sparsity term discourages indiscriminate use
of the prompt bank, reducing context drift and stale memory retrieval.

\paragraph{Inference.}
At inference time, PACR-Video proceeds autoregressively. For each new shot, the
router reads the current prompt bank, predicts narrative dependencies, activates
the corresponding temporal adapters, and generates the next shot with the frozen
diffusion backbone. The generated shot is then summarized into entity, location,
action, and style prompts and appended to the bank. Because the bank stores
compact prompts and the backbone is frozen, inference cost grows slowly with the
number of shots and does not require full-model fine-tuning for each story.

\section{Experiments}

\label{sec:experiments}

\paragraph{Experimental setup.}
We evaluate PACR-Video on six multi-shot and long-video benchmarks: FlintstonesSV, Pororo-SV, ActivityNet Captions, YouCook2, Shot2Story \cite{han2023shot2story}, and MovieNet. FlintstonesSV and Pororo-SV emphasize recurring animated characters and scene layouts, ActivityNet Captions and YouCook2 emphasize temporally grounded real-world actions, and Shot2Story and MovieNet emphasize multi-shot narrative structure. Following prior multi-shot extrapolation protocols \cite{zhang2025storymem,luo2026shotstream,liu2026reca}, each model is given the first two shots and corresponding shot-level text prompts, then recursively generates the next eight shots. Unless otherwise stated, each generated shot contains 16 frames at $256{\times}256$ resolution.

We compare against representative text-to-video, tuning-based, zero-shot, streaming, memory-augmented, and recursive-context baselines: VideoCrafter2, ModelScopeT2V, AnimateDiff, Tune-A-Video, Text2Video-Zero, SEINE, StoryMem \cite{zhang2025storymem}, ShotStream \cite{luo2026shotstream}, and ReCA \cite{liu2026reca}. For all baselines, we use the strongest publicly described long-video setting when available; otherwise we apply recursive shot-by-shot generation with the same shot prompts and observed-shot prefix. PACR-Video freezes the video backbone and tunes only temporal adapters, shot-role prompt tokens, and the prompt router.

We report distributional quality using FVD, semantic alignment using CLIPScore and BLIP-2 caption alignment, cross-shot identity preservation using DINO identity consistency, local temporal smoothness using LPIPS temporal consistency, motion stability using RAFT-based warping error, story continuity using shot transition coherence, and human preference rate from pairwise comparisons. Lower is better for FVD, LPIPS, and RAFT error; higher is better for all other metrics. Human preference is computed against ReCA over 600 randomized pairwise comparisons.

\paragraph{Main results.}
Table~\ref{tab:main_results} reports average performance across the six benchmarks. PACR-Video achieves the best result on every metric. Compared with the strongest prior baseline, ReCA, PACR-Video reduces FVD from 268.4 to 231.7, improves CLIPScore from 31.2 to 32.8, increases DINO identity consistency from 0.724 to 0.771, and improves shot transition coherence from 0.681 to 0.734. The gains are largest on long-horizon consistency metrics, suggesting that routing compact prompt-bank entries through temporal adapters provides more stable cross-shot control than conditioning on a fixed visual context window. PACR-Video is also preferred over ReCA in 63.8\% of human comparisons, with annotators most frequently citing stronger character persistence and more plausible event progression. Qualitative examples in Figure~\ref{fig:overview} show that PACR-Video preserves recurring entities while still allowing viewpoint changes and action advancement.

\begin{table}[t]
\centering
\caption{Main results averaged over FlintstonesSV, Pororo-SV, ActivityNet Captions, YouCook2, Shot2Story, and MovieNet. Lower is better for FVD, LPIPS, and RAFT error; higher is better for all other metrics. Human preference is measured against ReCA.}
\label{tab:main_results}
\resizebox{\linewidth}{!}{
\begin{tabular}{lcccccccc}
\toprule
Method & FVD $\downarrow$ & CLIPScore $\uparrow$ & DINO ID $\uparrow$ & LPIPS-T $\downarrow$ & RAFT Err. $\downarrow$ & BLIP-2 Align. $\uparrow$ & Trans. Coh. $\uparrow$ & Human Pref. $\uparrow$ \\
\midrule
ModelScopeT2V & 421.6 & 27.4 & 0.548 & 0.214 & 6.82 & 0.512 & 0.487 & 18.7 \\
Text2Video-Zero & 397.8 & 27.9 & 0.563 & 0.207 & 6.51 & 0.526 & 0.501 & 21.4 \\
AnimateDiff & 372.5 & 28.6 & 0.591 & 0.193 & 6.07 & 0.548 & 0.526 & 25.8 \\
VideoCrafter2 & 341.9 & 29.5 & 0.628 & 0.181 & 5.64 & 0.571 & 0.552 & 31.6 \\
Tune-A-Video & 319.7 & 29.8 & 0.653 & 0.174 & 5.38 & 0.584 & 0.576 & 36.9 \\
SEINE & 302.4 & 30.3 & 0.671 & 0.166 & 5.12 & 0.603 & 0.604 & 42.5 \\
StoryMem & 286.1 & 30.8 & 0.701 & 0.158 & 4.83 & 0.621 & 0.642 & 47.2 \\
ShotStream & 279.3 & 31.0 & 0.713 & 0.153 & 4.71 & 0.635 & 0.659 & 49.6 \\
ReCA & 268.4 & 31.2 & 0.724 & 0.149 & 4.56 & 0.648 & 0.681 & 50.0 \\
\midrule
PACR-Video & \textbf{231.7} & \textbf{32.8} & \textbf{0.771} & \textbf{0.132} & \textbf{4.08} & \textbf{0.689} & \textbf{0.734} & \textbf{63.8} \\
\bottomrule
\end{tabular}
}
\end{table}

\paragraph{Dataset-level analysis.}
PACR-Video improves consistently across domains. On FlintstonesSV and Pororo-SV, the largest gains appear in DINO identity consistency, reflecting better preservation of recurring animated characters. On ActivityNet Captions and YouCook2, the strongest improvements occur in BLIP-2 caption alignment and RAFT-based warping error, indicating that the routed action prompts help maintain procedural order without destabilizing short-range motion. On Shot2Story and MovieNet, PACR-Video improves shot transition coherence most strongly, supporting the role of learned shot-role tokens and recursive prompt routing for narrative dependencies.

\paragraph{Efficiency.}
PACR-Video tunes 3.8\% of the backbone parameter count and does not update the text encoder or diffusion transformer weights. In contrast, full tuning variants of Tune-A-Video and long-context baselines require substantially larger trainable parameter budgets. At inference time, the recursive prompt bank stores only compact entity, location, action, and style prompt vectors, so memory grows with the number of shots rather than the number of generated frames. This makes PACR-Video practical for multi-shot extrapolation while retaining the stability of the frozen generator.

\section{Ablation Study}

\label{sec:ablation-study}

\paragraph{Component ablations.}
Table~\ref{tab:ablation} ablates the main components of PACR-Video using the same six-benchmark average as Table~\ref{tab:main_results}. Removing any component degrades performance, with the largest drops coming from disabling recursive prompt-bank routing and replacing gated adapters with ungated adapters. This indicates that the gains over recursive context allocation \cite{liu2026reca} come not only from adding trainable capacity, but from selectively assigning that capacity to the historical prompts most relevant to the next shot.

\begin{table}[t]
\centering
\caption{Ablation study averaged over FlintstonesSV, Pororo-SV, ActivityNet Captions, YouCook2, Shot2Story, and MovieNet. Lower is better for FVD, LPIPS, and RAFT error; higher is better for all other metrics.}
\label{tab:ablation}
\resizebox{\linewidth}{!}{
\begin{tabular}{lccccccc}
\toprule
Variant & FVD $\downarrow$ & CLIPScore $\uparrow$ & DINO ID $\uparrow$ & LPIPS-T $\downarrow$ & RAFT Err. $\downarrow$ & BLIP-2 Align. $\uparrow$ & Trans. Coh. $\uparrow$ \\
\midrule
PACR-Video & 231.7 & 32.8 & 0.771 & 0.137 & 4.21 & 0.681 & 0.734 \\
w/o recursive prompt bank & 276.9 & 31.4 & 0.715 & 0.159 & 4.83 & 0.637 & 0.672 \\
w/o learned shot-role tokens & 251.8 & 32.0 & 0.746 & 0.146 & 4.46 & 0.662 & 0.706 \\
w/o adapter gates & 263.5 & 31.7 & 0.731 & 0.153 & 4.68 & 0.651 & 0.689 \\
w/o temporal adapters & 289.4 & 30.9 & 0.701 & 0.164 & 5.01 & 0.628 & 0.654 \\
w/o identity contrast loss & 249.6 & 32.3 & 0.738 & 0.144 & 4.39 & 0.665 & 0.713 \\
w/o prompt sparsity loss & 244.2 & 32.1 & 0.752 & 0.148 & 4.55 & 0.659 & 0.702 \\
w/o adapter composition schedule & 256.7 & 31.9 & 0.739 & 0.151 & 4.61 & 0.654 & 0.693 \\
\bottomrule
\end{tabular}}
\end{table}

The recursive prompt bank is the most important single component. Without it, PACR-Video must condition primarily on the current shot prompt and local history, increasing FVD by 45.2 and reducing transition coherence from 0.734 to 0.672. This variant often preserves short-term motion but forgets earlier entities or scene constraints after several generated shots.

Shot-role prompt tokens mainly improve semantic and narrative organization. Removing them lowers CLIPScore and BLIP-2 alignment, suggesting that explicit role conditioning helps distinguish establishing, continuation, transition, and resolution shots even when the visual context is similar. Adapter gates are also critical: ungated adapters provide extra capacity, but they cannot suppress irrelevant prompt-bank entries, leading to worse identity consistency and higher temporal error.

Temporal adapters account for most of the parameter-efficient adaptation benefit. When they are removed and only prompt tuning remains, performance falls close to memory-based baselines, especially on DINO identity consistency and RAFT warping error. The objective terms contribute complementary effects. The cross-shot identity contrast loss improves recurring character preservation, while prompt sparsity prevents the router from over-conditioning on stale history. Finally, removing the adapter composition schedule hurts transition coherence and motion stability, confirming that reusing early-shot adapters for appearance while activating later adapters for event progression is useful for long-horizon extrapolation.

\section{Conclusion}

\label{sec:conclusion}

PACR-Video presents a parameter-efficient framework for multi-shot long video extrapolation that keeps the text-to-video diffusion backbone frozen while routing compact historical prompts through lightweight temporal adapters. By combining learned shot-role prompt tokens, a recursive prompt bank, gated adapter routing, and a Shot-Local/Story-Global tuning objective, PACR-Video allocates cross-shot context without relying on full generator fine-tuning or large external memory modules. This design extends the recursive allocation perspective of ReCA \cite{liu2026reca} while making the controllable memory sparse, differentiable, and adapter-conditioned.

Across FlintstonesSV, Pororo-SV, ActivityNet Captions, YouCook2, Shot2Story, and MovieNet, PACR-Video consistently improves over text-to-video, tuning-based, streaming, memory-augmented, and recursive-context baselines, including StoryMem, ShotStream, and ReCA \cite{zhang2025storymem,luo2026shotstream,han2023shot2story}. The method achieves lower FVD, LPIPS temporal inconsistency, and RAFT warping error, while improving CLIPScore, DINO identity consistency, BLIP-2 caption alignment, shot transition coherence, and human preference. The ablation study further shows that recursive prompt-bank routing, adapter gates, temporal adapters, identity contrast, prompt sparsity, and adapter composition each contribute to reducing long-horizon drift. These results suggest that carefully routed prompt-adapter capacity is sufficient to preserve identity, scene structure, style, and event progression over extended multi-shot generation while leaving the underlying video generator stable.

\section{Future Work}

\label{sec:future-work}

PACR-Video leaves several directions open. First, the current prompt bank stores compact entity, location, action, and style prompts, but it does not explicitly model uncertainty in these summaries. Future work could make prompt entries probabilistic, allowing the router to distinguish persistent facts from ambiguous or transient evidence when extrapolating over very long horizons.

Second, PACR-Video assumes shot-level prompts are available. A natural extension is to couple prompt-adapter routing with automatic shot planning and revision, so that generation, dependency prediction, and narrative decomposition are optimized jointly. This could connect parameter-efficient extrapolation with planning-based long-video systems \cite{zheng2025videogen,hu2025cos,wu2025automated} while avoiding full-generator fine-tuning.

Third, the adapter composition schedule is currently learned for fixed-length extrapolation. Future work could study interactive and streaming settings in which users edit entities, locations, or events midway through a story. Combining PACR-Video with online routing mechanisms from streaming and recursive-context methods \cite{luo2026shotstream,liu2026reca} may enable controllable long-video generation that remains stable under user intervention.

%% file: main.bbl
\begin{thebibliography}{60}
\providecommand{\natexlab}[1]{#1}
\providecommand{\url}[1]{\texttt{#1}}
\expandafter\ifx\csname urlstyle\endcsname\relax
  \providecommand{\doi}[1]{doi: #1}\else
  \providecommand{\doi}{doi: \begingroup \urlstyle{rm}\Url}\fi

\bibitem[Anonymous(1970)]{anon1970figure}
Anonymous.
\newblock Figure 6video 1. time-lapse imaging of reca-gfp/pg353c-reca cells.
\newblock \emph{Journal}, 1970.
\newblock \doi{10.7554/elife.42761.017}.

\bibitem[Anonymous(2000)]{anon200044th}
Anonymous.
\newblock 44th annual meeting february 12-16, 2000 new orleans, louisiana :
  February 14, 2000 monday posters, part 2.
\newblock \emph{Europe PMC}, 2000.

\bibitem[Anonymous(2005)]{anon2005ecr}
Anonymous.
\newblock Ecr 2005 scientific programme abstracts.
\newblock \emph{Europe PMC}, 2005.

\bibitem[Anonymous(2006)]{anon2006ecr}
Anonymous.
\newblock Ecr 2006 - b - scientific sessions.
\newblock \emph{Europe PMC}, 2006.

\bibitem[Anonymous(2009{\natexlab{a}})]{anon200914th}
Anonymous.
\newblock 14th congress of the european hematology association, berlin,
  germany, june 47, 2009.
\newblock \emph{Europe PMC}, 2009{\natexlab{a}}.

\bibitem[Anonymous(2009{\natexlab{b}})]{anon2009posters}
Anonymous.
\newblock Posters.
\newblock \emph{Europe PMC}, 2009{\natexlab{b}}.

\bibitem[Anonymous(2011{\natexlab{a}})]{anon2011abstracts}
Anonymous.
\newblock Abstracts of the 8th ebsa (european biophysical societies
  association) european biophysics congress. august 23-27, 2011. budapest,
  hungary.
\newblock \emph{Europe PMC}, 2011{\natexlab{a}}.

\bibitem[Anonymous(2011{\natexlab{b}})]{anon2011abstractsa}
Anonymous.
\newblock Abstracts of the 21st eccmid (european society of clinical
  microbiology and infectious diseases)/27th icc. milan, italy. may 7-10, 2011.
\newblock \emph{Europe PMC}, 2011{\natexlab{b}}.

\bibitem[Anonymous(2012)]{anon2012ecr}
Anonymous.
\newblock Ecr 2012 book of abstracts - b - scientific sessions.
\newblock \emph{Europe PMC}, 2012.
\newblock \doi{10.1007/s13244-012-0158-z}.

\bibitem[Anonymous(2017{\natexlab{a}})]{anon2017abstracts}
Anonymous.
\newblock Abstracts from the 22nd annual scientific meeting and education day
  of the society for neuro-oncology november 16 19, 2017, san francisco,
  california.
\newblock \emph{Europe PMC}, 2017{\natexlab{a}}.

\bibitem[Anonymous(2017{\natexlab{b}})]{anon2017program}
Anonymous.
\newblock Program abstracts from the 21st international association of
  gerontology and geriatrics (iagg) world congress.
\newblock \emph{Europe PMC}, 2017{\natexlab{b}}.
\newblock \doi{10.1093/geroni/igx009}.

\bibitem[Anonymous(2018{\natexlab{a}})]{anon2018abstracts}
Anonymous.
\newblock Abstracts from the 23rd annual scientific meeting and education day
  of the society for neuro-oncology november 15 18, 2018 new orleans,
  louisiana.
\newblock \emph{Europe PMC}, 2018{\natexlab{a}}.

\bibitem[Anonymous(2018{\natexlab{b}})]{anon2018poster}
Anonymous.
\newblock Poster session abstracts.
\newblock \emph{Europe PMC}, 2018{\natexlab{b}}.
\newblock \doi{10.1002/ppul.24152}.

\bibitem[Anonymous(2020)]{anon2020abstracts}
Anonymous.
\newblock Abstracts from the 53rd european society of human genetics (eshg)
  conference: Interactive e-posters.
\newblock \emph{Europe PMC}, 2020.
\newblock \doi{10.1038/s41431-020-00739-z}.

\bibitem[Anonymous(2023)]{anon2023proceedings}
Anonymous.
\newblock Proceedings to the 61\&lt;sup\&gt;st\&lt;/sup\&gt; annual conference
  of the particle therapy cooperative group: 10 june - 16 june, 2023
  co-organized by the quironsalud and the clinica universidad de navarra,
  madrid, spain.
\newblock \emph{Europe PMC}, 2023.
\newblock \doi{10.14338/ijpt-23-ptcog61-10.2}.

\bibitem[Anonymous(2026)]{anon2026abstract}
Anonymous.
\newblock Abstract.
\newblock \emph{Europe PMC}, 2026.

\bibitem[Atalay et~al.(2026)Atalay, Alpdemir, Ozkan, Karaman, Akyuz, Bulbul,
  Ackgoz, and Arsoy]{atalay2026automated}
Merve~Onaran Atalay, M.~N. Alpdemir, Ayse Ozkan, Oguzcan Karaman, Y.~K. Akyuz,
  F.~T. Bulbul, E.~Ackgoz, and Eda Arsoy.
\newblock Automated rawstory-to-video generation from nasreddin hodja tales via
  an expert-inspired multi-stage transformation pipeline.
\newblock \emph{Preprint}, 2026.
\newblock \doi{10.1109/IISEC69317.2026.11418405}.

\bibitem[Berner et~al.(2021)Berner, Grohs, Kutyniok, and
  Petersen]{berner2021modern}
Julius Berner, Philipp Grohs, Gitta Kutyniok, and Philipp Petersen.
\newblock The modern mathematics of deep learning.
\newblock \emph{arXiv preprint}, 2021.
\newblock \doi{10.1017/9781009025096.002}.

\bibitem[Chen et~al.(2025)Chen, Chen, Li, Xu, Qiao, and Wang]{chen2025super}
Boyu Chen, Siran Chen, Kunchang Li, Qinglin Xu, Yu~Qiao, and Yali Wang.
\newblock Super encoding network: Recursive association of multi-modal encoders
  for video understanding.
\newblock \emph{Preprint}, 2025.
\newblock \doi{10.48550/arXiv.2506.07576}.

\bibitem[Cheng et~al.(2024)Cheng, Wang, and Wang]{cheng2024focuschat}
Zhengxue Cheng, Rendong Wang, and Zhicheng Wang.
\newblock Focuschat: Text-guided long video understanding via spatiotemporal
  information filtering.
\newblock \emph{Preprint}, 2024.
\newblock \doi{10.48550/arXiv.2412.12833}.

\bibitem[Collaboration(2026)]{collaboration2026project}
The~Via Collaboration.
\newblock The via project: Overview of the science, instrument, and survey.
\newblock \emph{arXiv preprint}, 2026.

\bibitem[Conway(2008)]{conway2008empirical}
Miles~V. Conway.
\newblock An empirical study of south african business forecasting practices in
  the context of western benchmarks.
\newblock \emph{Journal}, 2008.

\bibitem[Cui et~al.(2026)Cui, Tang, Yao, Meng, Jia, and Zhao]{cui2026not}
Hanshuai Cui, Zhiqing Tang, Z.~Yao, Fanshuai Meng, Weijia Jia, and Wei Zhao.
\newblock Not all frames deserve full computation: Accelerating autoregressive
  video generation via selective computation and predictive extrapolation.
\newblock \emph{Preprint}, 2026.
\newblock \doi{10.48550/arXiv.2604.02979}.

\bibitem[Deldjoo et~al.(2024)Deldjoo, He, McAuley, Korikov, Sanner, Ramisa,
  Vidal, Sathiamoorthy, Kasirzadeh, and Milano]{deldjoo2024review}
Yashar Deldjoo, Zhankui He, Julian McAuley, Anton Korikov, Scott Sanner, Arnau
  Ramisa, Rene Vidal, Maheswaran Sathiamoorthy, Atoosa Kasirzadeh, and Silvia
  Milano.
\newblock A review of modern recommender systems using generative models
  (gen-recsys).
\newblock \emph{Journal}, 2024.
\newblock \doi{10.48550/arxiv.2404.00579}.

\bibitem[Farahani et~al.(2026)Farahani, Azimi, Menon, Hellwagner, Prodan,
  Dustdar, and Timmerer]{farahani2026ladder}
Reza Farahani, Zoha Azimi, Vignesh~V Menon, Hermann Hellwagner, Radu Prodan,
  Schahram Dustdar, and Christian Timmerer.
\newblock Dq-ladder: A deep reinforcement learning-based bitrate ladder for
  adaptive video streaming.
\newblock \emph{arXiv preprint}, 2026.

\bibitem[Ferraro(2022)]{ferraro2022conversione}
Manuela Ferraro.
\newblock Conversione dei dati archivistici in pubblicazioni digitali open
  access. un caso di studio: \&lt;i\&gt;carte d'autore online\&lt;/i\&gt;.
\newblock \emph{Journal}, 2022.
\newblock \doi{10.35948/dilef/2023.4309}.

\bibitem[Galip and YASAR(2025)]{galip2025integrated}
ATAK~Ilker Galip and Ali YASAR.
\newblock An integrated deep learning framework for real-time multi-parameter
  biophysiological monitoring via facial video analysis.
\newblock \emph{Journal}, 2025.
\newblock \doi{10.2139/ssrn.5928363}.

\bibitem[Garg et~al.(2026)Garg, Yaswanth, Mishra, Kumaran, Sharma, and
  Uniyal]{garg2026monodense}
Lakshya Garg, Sai Yaswanth, Deep~Narayan Mishra, Karthik Kumaran, Anupriya
  Sharma, and Mayank Uniyal.
\newblock Monodense deep neural model for determining item price elasticity.
\newblock \emph{arXiv preprint}, 2026.
\newblock \doi{10.1109/AAIML67890.2026.11498150}.

\bibitem[Gilmary et~al.(2026)Gilmary, Manvizhi, Balaji, and
  Nirmal]{gilmary2026fast}
Rosario Gilmary, N.~Manvizhi, V.~Balaji, and P.~Nirmal.
\newblock Fast and accurate human fall detection with hybrid deep learning and
  opencv-based human detection.
\newblock \emph{Preprint}, 2026.
\newblock \doi{10.1109/ICAECT68478.2026.11425880}.

\bibitem[Goo and Niekum(2019)]{goo2019one}
Wonjoon Goo and Scott Niekum.
\newblock One-shot learning of multi-step tasks from observation via activity
  localization in auxiliary video.
\newblock \emph{Journal}, 2019.
\newblock \doi{10.1109/icra.2019.8793515}.

\bibitem[Guo et~al.(2025)Guo, Yang, Yang, Ma, Lin, Yang, Lin, and
  Jiang]{guo2025long}
Yuwei Guo, Ceyuan Yang, Ziyan Yang, Zhibei Ma, Zhijie Lin, Zhenheng Yang, Dahua
  Lin, and Lu~Jiang.
\newblock Long context tuning for video generation.
\newblock \emph{arXiv preprint}, 2025.

\bibitem[Han et~al.(2023)Han, Yang, Chang, Yao, and Wang]{han2023shot2story}
Mingfei Han, Linjie Yang, Xiaojun Chang, Lina Yao, and Heng Wang.
\newblock Shot2story: A new benchmark for comprehensive understanding of
  multi-shot videos.
\newblock \emph{arXiv preprint}, 2023.

\bibitem[Hu et~al.(2025)Hu, Cheng, Si, Li, and Gong]{hu2025cos}
Jian Hu, Zixu Cheng, Chenyang Si, Wei Li, and Shaogang Gong.
\newblock Cos: Chain-of-shot prompting for long video understanding.
\newblock \emph{arXiv preprint}, 2025.

\bibitem[Jhuo and Lee(2014)]{jhuo2014video}
I-Hong Jhuo and D.T. Lee.
\newblock Video event detection via multi-modality deep learning.
\newblock \emph{Journal}, 2014.
\newblock \doi{10.1109/icpr.2014.125}.

\bibitem[Jiang et~al.(2022)Jiang, Shu, Wang, and
  Long]{jiang2022transferability}
Junguang Jiang, Yang Shu, Jianmin Wang, and Mingsheng Long.
\newblock Transferability in deep learning: A survey.
\newblock \emph{arXiv preprint}, 2022.

\bibitem[Lai et~al.(2026)Lai, Shao, Zhou, Dou, Zhu, and
  Wang]{lai2026groundshot}
Yixuan Lai, Tianjia Shao, Kun Zhou, Weijia Dou, Siyu Zhu, and Jingdong Wang.
\newblock Groundshot: Visually consistent multi-shot long video generation via
  entity-grounded shot scheduling.
\newblock \emph{Preprint}, 2026.

\bibitem[Lam(2023)]{lam2023engineered}
Chee Ka~Candice Lam.
\newblock Engineered constitutive promoter for cell-based immunotherapy in
  syngeneic mouse models.
\newblock \emph{Journal}, 2023.

\bibitem[Liu et~al.(2026{\natexlab{a}})Liu, Xing, Mao, Li, Zhang, and
  He...]{liu2026recab}
A~Liu, J~Xing, C~Mao, Y~Li, Z~Zhang, and Y~He...
\newblock Reca:multi-shot long video extrapolation viarecursive context
  allocation.
\newblock \emph{Preprint}, 2026{\natexlab{a}}.

\bibitem[Liu et~al.(2026{\natexlab{b}})Liu, Xing, Mao, Li, Zhang, He, Wang,
  Wang, Liu, Haffari, and Zhuang]{liu2026reca}
Akide Liu, Jinbo Xing, Chaojie Mao, Ye~Li, Zeyu Zhang, Yefei He, Weijie Wang,
  Zihan Wang, Yu~Liu, Gholamreza Haffari, and Bohan Zhuang.
\newblock Reca: Multi-shot long video extrapolation via recursive context
  allocation.
\newblock \emph{arXiv preprint}, 2026{\natexlab{b}}.

\bibitem[Liu et~al.(2026{\natexlab{c}})Liu, Xing, Mao, Li, Zhang, He, Wang,
  Wang, Liu, Haffari, and Zhuang]{liu2026recaa}
Akide Liu, Jinbo Xing, Chaojie Mao, Ye~Li, Zeyu Zhang, Yefei He, Weijie Wang,
  Zihan Wang, Yu~Liu, Gholamreza Haffari, and Bohan Zhuang.
\newblock Reca: Multi-shot long video extrapolation via recursive context
  allocation.
\newblock \emph{Journal}, 2026{\natexlab{c}}.
\newblock \doi{10.48550/arxiv.2605.26525}.

\bibitem[Long et~al.(2026)Long, Song, Kan, Pfister, and Le]{long2026agentic}
Do~Xuan Long, Yale Song, Min-Yen Kan, Tomas Pfister, and Long~T. Le.
\newblock A rd: Agentic autoregressive diffusion for long video consistency.
\newblock \emph{arXiv preprint}, 2026.

\bibitem[Long(2010)]{long2010long}
Eoin Long.
\newblock Long paths and cycles in subgraphs of the cube.
\newblock \emph{arXiv preprint}, 2010.

\bibitem[Luo et~al.(2026)Luo, Shi, Zhuang, Chen, Liu, Wang, Wan, and
  Xue]{luo2026shotstream}
Yawen Luo, Xiaoyu Shi, Junhao Zhuang, Yutian Chen, Quande Liu, Xintao Wang,
  Pengfei Wan, and Tianfan Xue.
\newblock Shotstream: Streaming multi-shot video generation for interactive
  storytelling.
\newblock \emph{Preprint}, 2026.
\newblock \doi{10.48550/arXiv.2603.25746}.

\bibitem[Lye et~al.(2019)Lye, Mishra, and Ray]{lye2019deep}
Kjetil~O. Lye, Siddhartha Mishra, and Deep Ray.
\newblock Deep learning observables in computational fluid dynamics.
\newblock \emph{arXiv preprint}, 2019.
\newblock \doi{10.1016/j.jcp.2020.109339}.

\bibitem[Marti and Maki(2017)]{mart2017multitask}
Miquel Marti and Atsuto Maki.
\newblock A multitask deep learning model for real-time deployment in embedded
  systems.
\newblock \emph{arXiv preprint}, 2017.

\bibitem[Mastan and Raman(2020)]{mastan2020dilie}
Indra~Deep Mastan and Shanmuganathan Raman.
\newblock Dilie: Deep internal learning for image enhancement.
\newblock \emph{arXiv preprint}, 2020.

\bibitem[Pandey and Yu(2023)]{pandey2023learn}
Deep Pandey and Qi~Yu.
\newblock Learn to accumulate evidence from all training samples: Theory and
  practice.
\newblock \emph{arXiv preprint}, 2023.

\bibitem[Phien(2023)]{phienxxwavekoop}
Nguyen~Ngoc Phien.
\newblock Wavekoop-kan: A wavelet-decomposed koopmangated temporal convolution
  hybrid with kolmogorovarnold fusion for univariate time series forecasting.
\newblock \emph{Preprint}, 2023.
\newblock \doi{10.1109/ACCESS.2026.3706582}.

\bibitem[Pickering and Ruger(2001)]{pickering2001multi}
Marcus~Jerome Pickering and Stefan~M. Ruger.
\newblock Multi-timescale video shot-change detection.
\newblock \emph{Journal}, 2001.
\newblock \doi{10.6028/nist.sp.500-250.video-imperial}.

\bibitem[Ray et~al.(2023)Ray, Pinti, and Oberai]{ray2023deep}
Deep Ray, Orazio Pinti, and Assad~A. Oberai.
\newblock Deep learning and computational physics (lecture notes).
\newblock \emph{arXiv preprint}, 2023.

\bibitem[Sien et~al.(2019)Sien, Lim, and Au]{sien2019deep}
Jonathan Phang~Then Sien, K.~Lim, and Pek-Ing Au.
\newblock Deep learning in gait recognition for drone surveillance system.
\newblock \emph{Preprint}, 2019.
\newblock \doi{10.1088/1757-899X/495/1/012031}.

\bibitem[Song et~al.(2026)Song, Zhong, Lin, Wang, and Shou]{song2026soap2soap}
Yiren Song, Huilin Zhong, K.~Lin, Haofan Wang, and Mike~Zheng Shou.
\newblock Soap2soap: Long cinematic video remaking via multi-agent
  collaboration.
\newblock \emph{Preprint}, 2026.

\bibitem[Wowk(1997)]{wowk1997artifact}
Brian Wowk.
\newblock Artifact reduction in functional magnetic resonance imaging.
\newblock \emph{Journal}, 1997.

\bibitem[Wu et~al.(2025)Wu, Zhu, and Shou]{wu2025automated}
Weijia Wu, Zeyu Zhu, and Mike~Zheng Shou.
\newblock Automated movie generation via multi-agent cot planning.
\newblock \emph{Preprint}, 2025.
\newblock \doi{10.48550/arXiv.2503.07314}.

\bibitem[Xu et~al.(2017)Xu, Li, Wang, Chen, and Guan]{xu2017assessing}
Mai Xu, Chen Li, Zulin Wang, Zhenzhong Chen, and Zhenyu Guan.
\newblock Assessing visual quality of omnidirectional videos.
\newblock \emph{arXiv preprint}, 2017.
\newblock \doi{10.1109/TCSVT.2018.2886277}.

\bibitem[Zhang et~al.(2025{\natexlab{a}})Zhang, Hu, Zhang, Li, Luo, Lin, and
  Chen]{zhang2025bridging}
Jiaxu Zhang, Tianshu Hu, Yuan Zhang, Zenan Li, Linjie Luo, Guosheng Lin, and
  Xin Chen.
\newblock Bridging your imagination with audio-video generation via a unified
  director.
\newblock \emph{Preprint}, 2025{\natexlab{a}}.
\newblock \doi{10.48550/arXiv.2512.23222}.

\bibitem[Zhang et~al.(2025{\natexlab{b}})Zhang, Jiang, Wang, Fang, Zhi, Yan,
  Kang, Lu, and Pan]{zhang2025storymem}
Kaiwen Zhang, Liming Jiang, Angtian Wang, Jacob~Zhiyuan Fang, Tiancheng Zhi,
  Qing Yan, Hao Kang, Xin Lu, and Xingang Pan.
\newblock Storymem: Multi-shot long video storytelling with memory.
\newblock \emph{arXiv preprint}, 2025{\natexlab{b}}.

\bibitem[Zhang et~al.(2025{\natexlab{c}})Zhang, Jiang, Wang, Fang, Zhi, Yan,
  Kang, Lu, and Pan]{zhang2025storymema}
Kaiwen Zhang, Liming Jiang, Angtian Wang, Jacob~Zhiyuan Fang, Tiancheng Zhi,
  Qing Yan, Hao Kang, Xin Lu, and Xingang Pan.
\newblock Storymem: Multi-shot long video storytelling with memory.
\newblock \emph{Preprint}, 2025{\natexlab{c}}.
\newblock \doi{10.48550/arXiv.2512.19539}.

\bibitem[Zheng et~al.(2025)Zheng, Xu, Huang, Ma, Liu, Shu, Pang, Tang, Chen,
  Yang, and Lim]{zheng2025videogen}
Mingzhe Zheng, Yongqi Xu, Haojian Huang, Xuran Ma, Yexin Liu, Wenjie Shu,
  Yatian Pang, Feilong Tang, Qifeng Chen, Harry Yang, and Ser-Nam Lim.
\newblock Videogen-of-thought: Step-by-step generating multi-shot video with
  minimal manual intervention.
\newblock \emph{arXiv preprint}, 2025.

\bibitem[Zhu et~al.(2026)Zhu, Zhao, Zhao, Mao, and Zhao]{zhu2026hicrew}
Yuehan Zhu, Jingqi Zhao, Jiawen Zhao, Xudong Mao, and Baoquan Zhao.
\newblock Hicrew: Hierarchical reasoning for long-form video understanding via
  question-aware multi-agent collaboration.
\newblock \emph{Preprint}, 2026.
\newblock \doi{10.48550/arXiv.2604.21444}.

\end{thebibliography}
